\documentclass[a4paper]{article}
\usepackage[preprint,nonatbib]{nips_2018} 

\usepackage[english]{babel}

\usepackage[utf8]{inputenc} 
\usepackage[T1]{fontenc}    
\usepackage{url}            
\usepackage{booktabs}       
\usepackage{amsfonts}       
\usepackage{nicefrac}       
\usepackage{microtype}      
\usepackage{subcaption}
\usepackage{caption} 
\usepackage{amsmath}
\usepackage{graphicx}
\usepackage[colorinlistoftodos]{todonotes}
\usepackage{tikz}
\usetikzlibrary{decorations.pathreplacing}
\usetikzlibrary{shapes.multipart}
\usetikzlibrary{arrows.meta}

\tikzstyle{myPath} = [-{Latex[length=2mm,width=2mm]}, line width=0.4mm]
\usepackage{algpseudocode}
\usepackage[ruled]{algorithm2e}

\definecolor{niceblue}{HTML}{0074D9}
\usepackage{marginnote}
\DeclareMathOperator*{\argmax}{arg\,max}

\title{Are You Sure You Want To Do That?\\ Classification with Verification}

\newcommand*\samethanks[1][\value{footnote}]{\footnotemark[#1]}
\author{
  Harris Chan\thanks{Department of Computer Science at
  University of Toronto, Vector Institute}\\
  \texttt{hchan@cs.toronto.edu} \\
  \And
  Atef Chaudhury\samethanks\\
  \texttt{atef@cs.toronto.edu}
  \And
  Kevin Shen\samethanks\\
  \texttt{shenkev@cs.toronto.edu}
}

\begin{document}
\maketitle

\begin{abstract}
  Classification systems typically act in isolation, meaning they are required to implicitly memorize the characteristics of all candidate classes in order to classify. The cost of this is increased memory usage and poor sample efficiency. We propose a model which instead verifies using reference images during the classification process, reducing the burden of memorization. The model uses iterative non-differentiable queries in order to classify an image. We demonstrate that such a model is feasible to train and can match baseline accuracy while being more parameter efficient. However, we show that finding the correct balance between image recognition and verification is essential to pushing the model towards desired behavior, suggesting that a pipeline of recognition followed by verification is a more promising approach. 
\end{abstract}

\section{Introduction}

Supervised classification is one of the most common problems in machine-learning, and is often addressed with a wholly recognition based approach \cite{rawat2017deep}. Systems are expected to have memory, often implicit, of all possible candidate classes, and when an example is provided to the system, it is expected to leverage this memory in order to determine its class. However, this behavior may not always be suitable, in particular for cases in which there are a large number of candidate classes, or in situations with limited data. Consider a human tasked with classifying an uncommon breed of dog. A common action in this case to form a hypothesis about the breed, and then consult reference images for that breed to verify the hypothesis. This allows for more accurate classification and also relaxes the requirement of retaining high-fidelity memories of all classes. In this paper, we propose a framework, called Recognition-Verification Neural Network (RVNN), which uses this type of behavior to aid a neural network with classification. By reducing the memory requirement associated with wholly recognition based classification, we aim to show reductions in implicit memory (as measured by number of parameters), and better data-efficiency in training.


\begin{figure}[h]
\centering
\begin{subfigure}[b]{0.2\textwidth}
\centering
\includegraphics[width=0.4\linewidth,keepaspectratio]{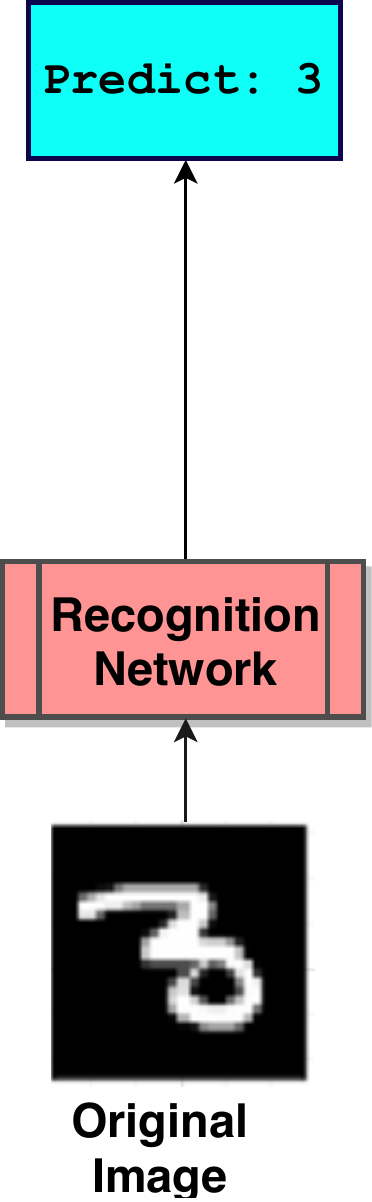}
\caption{Recognition}
\label{fig:model_recognition}
\end{subfigure} \hspace{5mm}
\begin{subfigure}[b]{0.24\textwidth}
\includegraphics[width=0.9\linewidth,keepaspectratio]{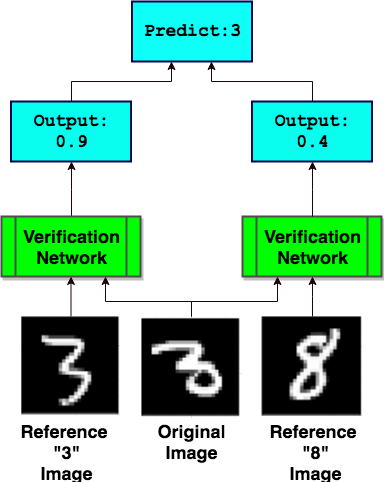}
\caption{Verification}
\label{fig:model_verification}
\end{subfigure} \hspace{5mm}
\begin{subfigure}[b]{0.4\textwidth}
\includegraphics[width=0.9\linewidth,keepaspectratio]{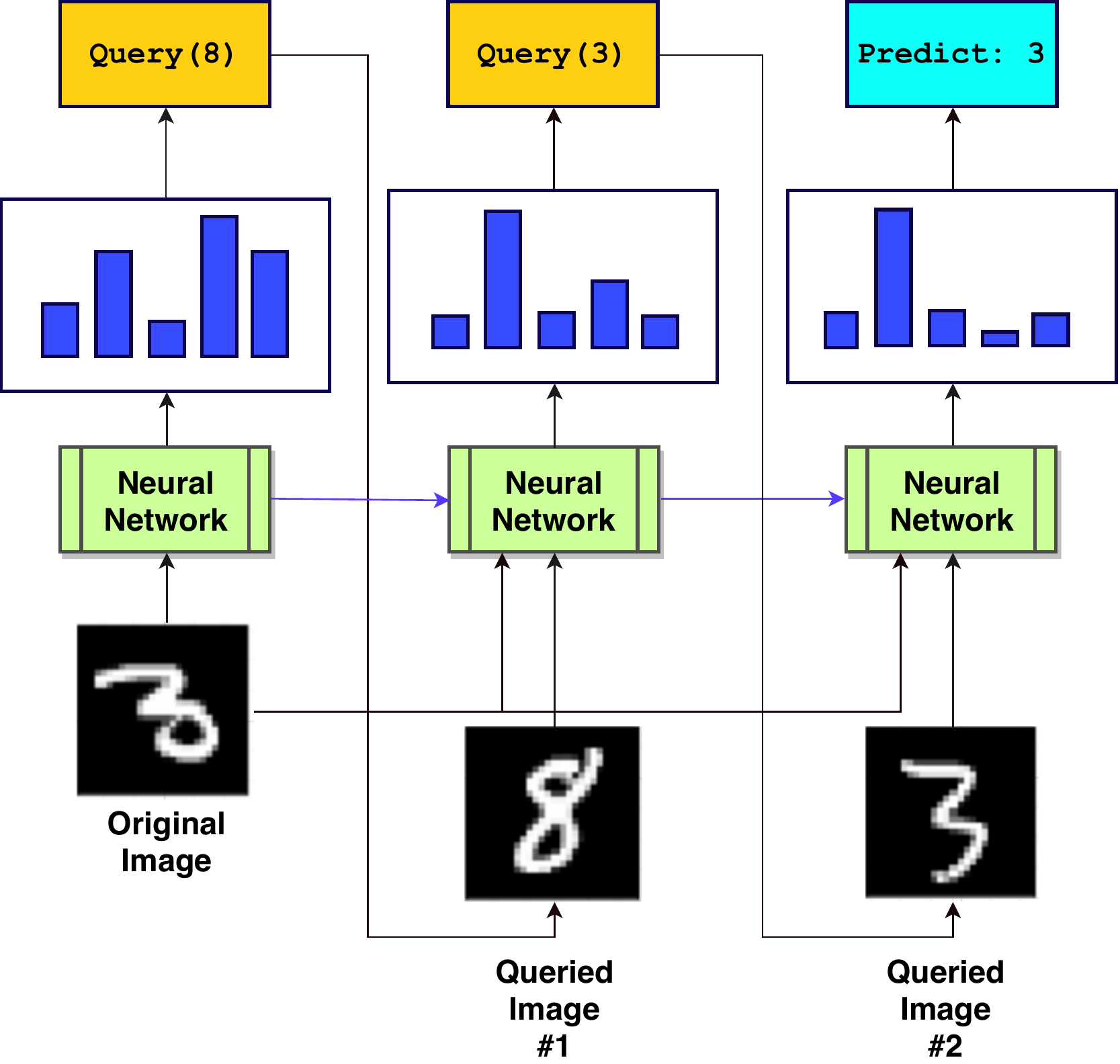}
\caption{Recognition Verification Neural Network (RVNN)}
\label{fig:main}
\end{subfigure}
\caption{Overview of our hybrid model in contrast with two opposing approaches. (a) Recognition network directly predicts the class given the input. (b) Verification network predicts binary output indicating the amount of similarity or likelihood that they belong in the same class. The verification network can be used to compare to all reference images from each class to produce the final class prediction. (c) Our approach, RVNN, queries for reference image from a particular class at each time step, and makes a class prediction at the last time step. \label{fig:overview}}
\end{figure}

Notably, our system uses non-differentiable queries for reference images in order to assist with its classification task. As shown in Figure. \ref{fig:overview}, our model is designed to iteratively query for support images in order to perform its classification task. Then at each subsequent time-step the model is given the image from its prior query, which it can compare with the task image in order to refine its hypothesis, before finally producing a prediction. 

\section{Related Works}

\subsection{Verification Based Classification and Few-Shot Learning}

While recognition based classification is most common, verification based classification has been explored, primarily in the domain of few-shot classification. Siamese Networks use two Convolutional Neural Networks (CNN) with shared weights to compare if two images are from the same class \cite{koch2015siamese}. Then to classify, the image is pair-wise compared with a support image from every class and image with the maximum similarity score is chosen. Matching networks extends the verification-based approach by outputting a prediction based on a weighted-sum of similarity across classes \cite{vinyals2016matching}. Additionally the work introduces an episodic-training regime which encourages the model to better learn for the one-shot learning scenario. Prototypical Networks uses Euclidean Distance in embedding space as a verification metric rather than a learned metric, while maintaining the same training regime as Matching Networks to encourage different classes to have distant means in embedding space \cite{snell2017prototypical}. One recent work outside of few-shot learning domain is the Retrieval-Augmented Convolutional Neural Networks (RaCNN) \cite{RaCNN}, which combines CNN recognition network with a retrieval engine for support images to help increase adversarial robustness.

For all the above few shot learning approaches, verification with support images from all classes are required before a classification decision is made. Hence the classification decision is solely derived from verifications. RaCNN is closer to our approach, which uses a hybrid between verification and recognition. However, RaCNN simply retrieves the $K$ closest support image neighbours in the embedding space, whereas our model is required to form a hypothesis as to which class to compare with. In cases in which there are a large number of classes, we expect our approach to excel. As well, this introduces a non-differentiable component in our model not present in previous work.

\subsection{Consulting External Knowledge Bases}

Prior work has also looked at the concept of incorporating external knowledge into the decision making process of neural networks, often with non-differentiable components. Buck et al. learn to formulate queries for a search engine in order to answer trivia questions \cite{buck2017ask}. The model must handle the non-differentiable component of interacting with an external environment, in this case a search engine. Another common source of external knowledge is a human, i.e. the human in the loop approach. For example, Ling et al. use non-differentiable instructions from a human teacher to better caption images and Thomaz et al. do the same for teaching an agent to navigate households \cite{ling2017teaching} \cite{thomaz2006reinforcement}.

Our paper shares a similarity in method to these works, as the model queries for external knowledge, in this case support images, in order to achieve its primary goal of classification. Notably, our model is also required to perform multiple non-differentiable queries rather than a single non-differentiable step. As well, while tasks such as question-answering and captioning have used this approach, this paper introduces the approach to the setting of classification.

\subsection{Application of Gradient Estimators}

When optimizing non-differentiable components, gradients are often estimated with the REINFORCE algorithm \cite{williams1992simple}. However, due to the high variance of the method, training complex models is often time-consuming or not feasible. Recently, several new gradient estimators have been proposed to address this issue. The Gumbel-Softmax approach injects Gumbel noise to form a continuous relaxation of a discrete choice \cite{jang2016categorical}. While, this approach biases the gradients, it has shown good empirical results. For this work we use this method in order to generate non-differentiable queries for support images.  


\section{Model Description}
\begin{figure}[h]
\centering
\includegraphics[width=0.85\textwidth,height=8cm,keepaspectratio]{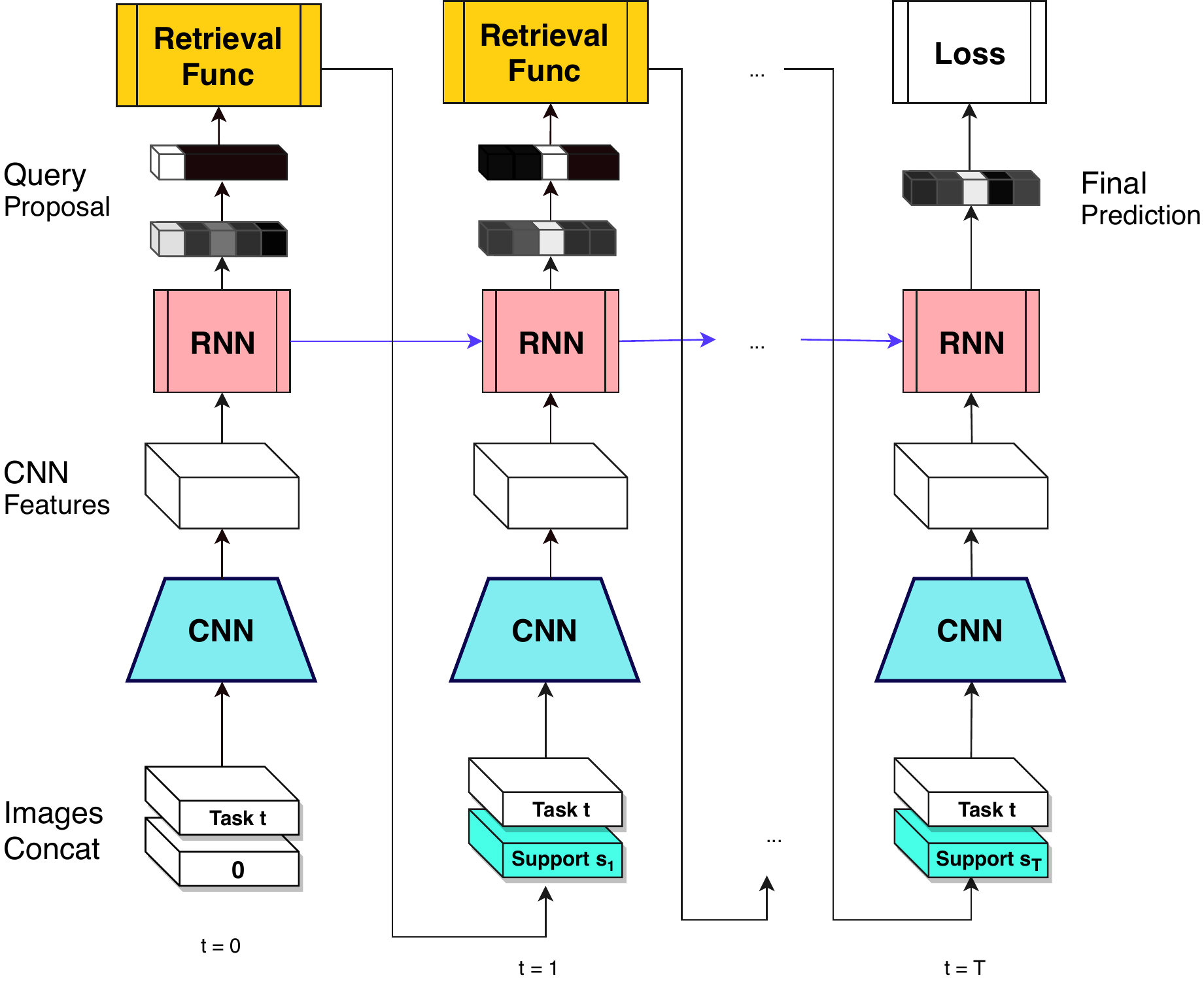}
\caption{\label{fig:model} Diagram of model architecture. At each time-step the original image along with the queried image are passed into the CNN module. The output of this is passed into the RNN which then outputs a query. This occurs for a fixed number of time-steps until the model outputs a prediction.}
\end{figure}

The model consists of three key components. First is a CNN-based module $f_{cnn}$ tasked with both recognition and verification. The module takes as input the task image $t$ along with a support image $s$, and then outputs some feature vector $v$. This feature vector is then input to a recurrent module $f_{rnn}$ tasked with tracking hypotheses and performing the high-level querying logic. The hidden state of the RNN $h$ is then input to the final querying component $f_q$ which converts it to a discrete query $q$, and then returns a new support image corresponding to that query. The model is run for a fixed number of time steps $N$, and at the last time step the the hidden state of the RNN is fed into a function $f_p$ in order to produce a prediction $C$ for the class. Figure \ref{fig:model} illustrates the full process.

\begin{algorithm}[H]
	\SetAlgoLined
	\DontPrintSemicolon
    \SetSideCommentRight
	\caption{Classification with References}
	\KwIn{Task image $\mathbf{T}$ to be classified}
	\KwResult{Predicted class $\mathbf{C}$ for the task image} 
    $\mathbf{S_1} \leftarrow 0$ $\mathbf{h_0} \leftarrow 0$  \quad \tcp{Initialize support image and RNN state}
	\For{$\mathbf{n}$ in $\mathbf{1 ... N}$} {
        $\mathbf{v_n}$ = $f_{cnn}(\mathbf{T}, \mathbf{S_n})$\;
        $\mathbf{h_n}$ = $f_{rnn}(\mathbf{v_n}, \mathbf{h_{n-1}})$\;
        $\mathbf{S_{n+1}}$ = $f_{q}(\mathbf{h_n})$\;		
	}
    $\mathbf{C} = f_p(\mathbf{h_N})$\;
	\Return{$\mathbf{C}$}
\end{algorithm}

The subsequent sections detail the implementations of the three components as well as training considerations.

\subsection{Verification Architecture $f_{cnn}$} \label{f_cnn}
\begin{figure}[h]
\centering
\begin{subfigure}[b]{0.3\textwidth}
\centering
\includegraphics[width=0.4\linewidth,keepaspectratio]{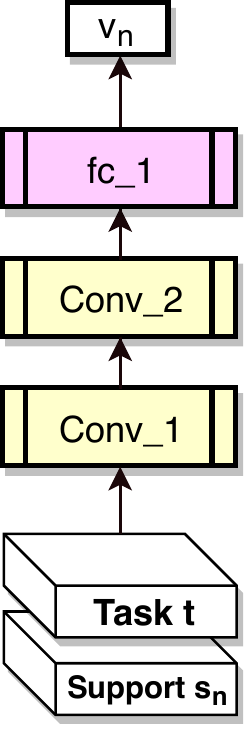}
\caption{Beginning Concatenate}
\label{fig:cnn_concatbegin}
\end{subfigure}
\begin{subfigure}[b]{0.3\textwidth}
\includegraphics[width=0.9\linewidth,keepaspectratio]{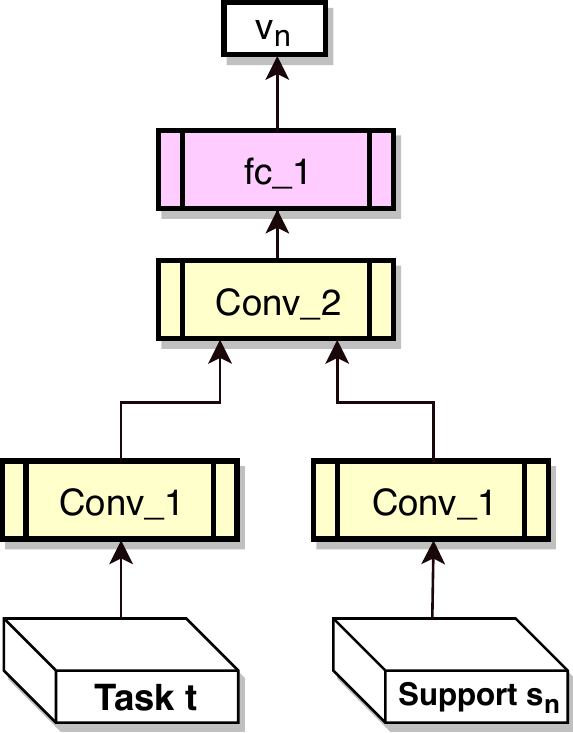}
\caption{Middle Concatenate}
\label{fig:cnn_concatmiddle}
\end{subfigure}
\begin{subfigure}[b]{0.3\textwidth}
\includegraphics[width=0.9\linewidth,keepaspectratio]{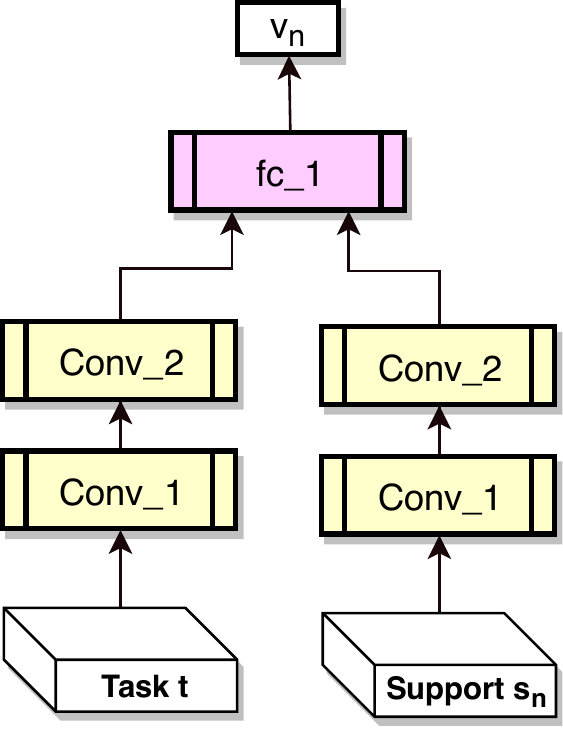}
\caption{End Concatenate}
\label{fig:cnn_concatend}
\end{subfigure}
\caption{Various architecture choices for the CNN Verification Network $f_{cnn}$ \label{fig:cnn_arch}}
\end{figure}

We explored three possible implementations of the CNN-based module $f_{cnn}$, which vary according to the layer at which information between the task image $t$ and support image $s$ are concatenated. The variants are illustrated in Figure \ref{fig:cnn_arch}. In general the architecture uses 2 convolutional layers and one fully connected layer. In the "Beginning Concatenate" (Fig. \ref{fig:cnn_concatbegin}), the task image and support image are concatenated along the channel dimension, then passed into the convolutional layers. In the "Middle Concatenate" (Fig. \ref{fig:cnn_concatmiddle}), the output feature maps from the first convolution layer from the task and support images are concatenated channel-wise. 
Similarity, the "End Concatenate" (Fig. \ref{fig:cnn_concatend}) concatenates the outputs from the second convolutional layers channel-wise. Note that the weights are tied between the corresponding convolution layers for the original and query images. 

\subsection{Recurrent Querying Model $f_{rnn}$}
The recurrent querying model $f_{rnn}$ was implemented using a Gated Recurrent Unit (GRU) \cite{gru}. We also considered passing in additional information to $f_{rnn}$, such as the query that was used in the previous time steps. 

\subsection{Implementing $f_q$ for Training}
We implemented $f_q$ as sampling a class based on the categorical probability given by the softmax of the logits from the $f_{rnn}$. Therefore, $f_q$ can be written as:
\begin{align}
\mathbf{p_n} &= \text{softmax}(\mathbf{h_{n}})\\
f_q(\mathbf{h_{n}}) &= \mathbf{S_{n+1}} \sim \text{Categorical}(\mathbf{p_n})
\end{align}

To sample $\mathbf{S_{n+1}}$, we can use the Gumbel-Max trick \cite{lucegumbelmax, gumbelmaxmaddison}:
\begin{align}
f_q(\mathbf{h_{n}}) &= \text{one\_hot}\bigg(\argmax_i [\mathbf{g}_i + \log \mathbf{p}_i ]\bigg)
\end{align}

where $\mathbf{g}_i ... \mathbf{g}_k$ are i.i.d samples drawn from $\text{Gumbel}(0,1)$ distribution. However, the $\argmax$ operator is not differentiable, so instead we explored two approaches during training. 

The first is to use the Gumbel-Softmax trick, also known as the Concrete estimator \cite{gumbelsoftmax, concrete}. We relaxed the $\argmax$ operator to a differentiable softmax function with temperature parameter $\tau$:

\begin{align}
\mathbf{S_n}_i &= \frac{\exp((\mathbf{g}_i + \log \mathbf{p}_i)/\tau)}{\sum^k_{j=1} \exp((\mathbf{g}_j + \log \mathbf{p}_j)/\tau)} \quad \text{for} \ i\ =\  1,\ldots,k
\end{align}

The $\tau$ parameter is annealed exponentially from $\tau = 1$ to $\tau=0.5$ as the training iterations progresses. 

The second approach is to use simple Straight-Through estimator \cite{straightthrough}. In the forward pass, we apply the Gumbel-Max trick to take discrete query choices. Then on the backward pass, we set the derivative of the query with respect to the softmax probabilities to be \textit{identity} so that the out-going gradient from the $\argmax$ operator is equal to the incoming gradient during backpropagation:

\begin{align}
\frac{\partial{\mathcal{L}}}{\partial \mathbf{p}} &= \frac{\partial{\mathcal{L}}}{\partial \mathbf{S}}\frac{\partial{\mathbf{S}}}{\partial \mathbf{p}} = \frac{\partial{\mathcal{L}}}{\partial \mathbf{S}} 
\end{align}

\subsection{Comparison to Existing Work}
We highlight that our model differ from several existing networks in various aspects. In the models for few-shot learning, such as Matching Networks and Prototypical Networks, their approach is similar to the verification approach which performs verification between the input image and the support images from \textit{all} classes. In the case of Matching Networks, they use cosine similarity between the embedding, while Prototypical networks use Euclidean distance as a measure of similarity. Our model aims to \textit{not} compare support images from all classes, but rather iteratively query for the most promising class's support images. We believe that this approach will be able to scale to larger number of classes. 

In comparison to RaCNN, we use a different retrieval engine and have the notion of memory (via the recurrent network $f_{rnn}$. RaCNN retrieves from its support set the $K$ nearest neighbour to the input image in the embedding space (i.e. output from pretrained CNN feature extractor), and produces a combined single vector representation with respect to the input image using attention mechanism. In contrast, our query function samples an image given the class probabilities, which can viewed as a form of hard attention on a particular support image. The use of recurrent network to perform the next query can then be interpreted as a recurrent attention over the support set. RaCNN has only a \textit{single} non-differentiable query, while ours perform \textit{multiple} non differentiable queries. 

\section{Experiments}

We perform experiments to assess both the overall performance of the model and to better understand its behavior. Based on our hypotheses, overall performance is judged via reduced parameter usage and sample efficiency. With regards to behavior, we focus on understanding the policy which the model learns, and how it can be influenced. To do so we run several ablated versions of our model, isolating the effect of each component. We also investigate the effects of decomposing the CNN-module into recognition and verification components, and the effects that they have on model behavior. All tests are conducted on the MNIST digit-classification dataset \cite{LeCun}.

\subsection{Parameter and Sample Efficiency}

The performance of the model is assessed by both reduced parameter usage and sample efficiency. Reduced parameter usage is measured relative to a baseline model, in this case the CNN architecture from \cite{PyTorchMNIST}. We look to see whether an identical level of accuracy can be achieved by our model, but with fewer parameters than the baseline. Sample efficiency is measured by taking models that achieved similar levels of accuracy by training on the full dataset, and then training them on a subset of the data and recording the loss in test accuracy. The model with a lesser reduction in accuracy would be considered more sample efficient. We test both smaller and larger versions of our model (as measured by number of parameters), against smaller and larger versions of the baseline model.

\subsection{Query Result Modification}

The performance of the model alone does not indicate whether our approach is functioning as intended. It is possible that the model may simply use the RNN as additional computational resources and ignore all information from the queries. To test this, we experiment with modifying the query results during inference and observe their effect on model performance. We test supplying blank information or   incorrect images as query results during inference. If the model is using the query information in a valuable way, we expect this to significantly harm model accuracy. 

\subsection{Architectural Considerations and Hyper-parameters}

We also experiment with several small modifications to our architecture as well as a few hyper-parameters that are unique to our model. We list them here below.

\begin{itemize}
  \item Architectural Considerations 
  \begin{itemize}
    \item Query Memory (QM): The query from the past time step is passed to the RNN.
    \item Weighted CNN Output (WC): In the case of non-straight-through gumbel a weighted-sum is required. This is either done in pixel space or in latent space (after the CNN module).
    \item Separate RNN Heads (SH): The RNN output is split into two components, one for predicting and the other for querying, or both actions are derived from the same head.
    \item CNN Module : The use of Concat Begin, Middle or End as defined in section \ref{f_cnn}
  \end{itemize}
  \item Hyper-parameters
  \begin{itemize}
    \item Size of CNN as measured by number of channels
    \item Size of RNN as measured by hidden size
    \item Gumbel-Softmax anneal rate, and use of straight-through
    \item Number of queries made by the RNN
  \end{itemize}
\end{itemize}

\subsection{Decomposed CNN Modules and Query Policies}

Our model assumes that the CNN module will perform some hybrid form of recognition and verification, and that its output will be a unified representation of this computation to be provided to the RNN. In order to understand the impacts of the module's ability to recognize and compare at a granular level, we also experiment with a decomposed version of the CNN module consisting of a pre-trained classifer and comparator. By varying the strength of these we are able to see how the RNN learns to adapt its policy to varying levels of accuracy, and see whether a better result can be achieved when they are used in tandem. We also experiment with fixing query policy to assess whether the RNN actually learns intelligent query behavior.  

\begin{figure}[h]
\centering
\includegraphics[width=0.65\textwidth,height=6.5cm,keepaspectratio]{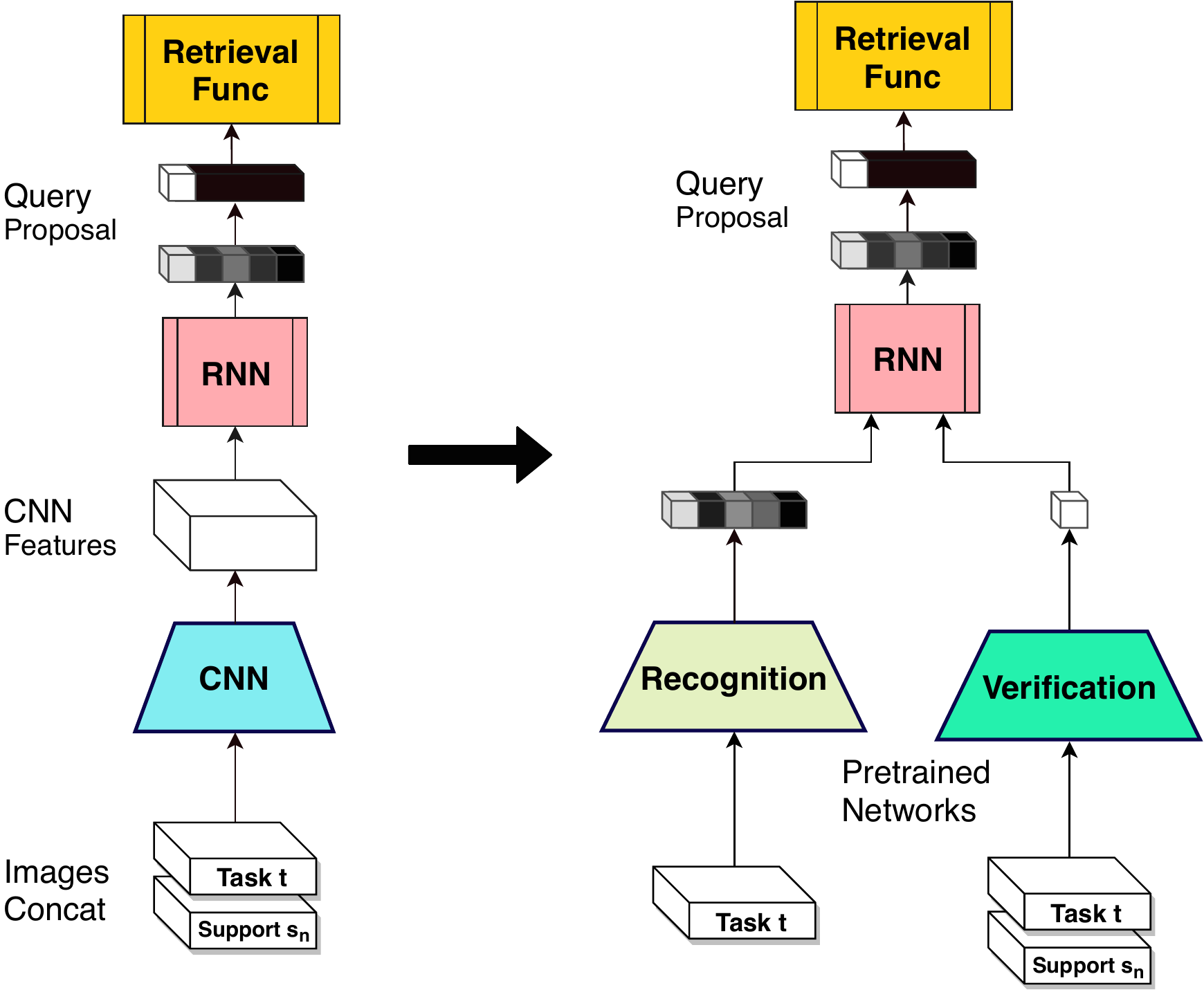}
\caption{Decomposing CNN module into explicit pre-trained recognition and verification networks. \label{fig:decompose_arch}}
\end{figure}

\section{Results}

\subsection{Parameter and Sample Efficiency}

\begin{table}[h!]
\begin{center}
 \begin{tabular}{||c | c | c | c | c ||} 
 \hline
 Smaller Models & Acc (\%) & Parameters (thousands) & 10\%-Data Acc (\%) & 1\%-Data Acc (\%) \\ 
 \hline
 Baseline & 99.09 & 27.7 & \textbf{97.94} & \textbf{93.01} \\ 
 \hline
 Ours & 99.09 & \textbf{13.9} & 97.73 & 92.26 \\ 
 \hline\hline
 Larger Models & Acc (\%) & Parameters (thousands) & 10\%-Data Acc (\%) & 1\%-Data Acc (\%) \\
 \hline
 Baseline & 99.29 & 53.2 & \textbf{98.39} & \textbf{94.90} \\ 
 \hline
 Ours & 99.29 & \textbf{34.2} & 98.12 & 92.89 \\  
 \hline
\end{tabular}
\caption{Parameter usage and sample efficiency for baseline model vs ours. Best versions of our model are reported against best versions of baseline}
\label{table:perf}
\end{center}
\end{table}
From our overall performance metrics we observe that at both smaller and larger sizes of model, our architecture achieves the same accuracy as the baseline but with approximately half the parameter usage. This is in agreement with our hypotheses that our model would be more parameter efficient. However, with regards to sample efficiency, our model performs worse than the baseline, going against our initial hypotheses. This suggests that our model may not have learned the behavior we expected it to. 

The results of the architecture/hyper-parameter search support this conclusion as well. Performance of the model was largely unaffected by parameters related to querying, such as gumbel-temperature, anneal rates and number of queries. Architectural modifications such as query memory or separate heads also had little effect on performance. The only key varying components were the CNN and RNN sizes, of which, the best model (for a fixed number of parameters) had the largest possible CNN with the smallest possible RNN. As the model is unaffected by parameters related to querying it appears as though the model does not rely on queries to classify, and instead works as a standard classifier (hence the performance gains from a large CNN and small RNN). The underlying behavior of the model is discussed further in the next sections.

\subsection{Query Result Modification}

\begin{table}[h!]
\begin{center}
 \begin{tabular}{|| c | c ||} 
 \hline
 \textbf{Query Result} & \textbf{Accuracy (\%)} \\ 
 \hline
 Standard & 99.29 \\ 
 \hline
 Blank & 97.44 \\ 
 \hline
 Mistaken & 98.02 \\ 
 \hline
\end{tabular}
\caption{Inference accuracy for modified query results}
\label{table:behav}
\end{center}
\end{table}
Replacing the queries with either blank or incorrect queries had limited effect on the model's performance. This supports the notion that the model is not actually using the queries to classify. Instead it is simply acting as a classifier with an RNN component. This is also the likely reason for the reduced parameter usage. As the model has the ability to do multiple iterations of computation via the RNN, it is possible that this results in a trade-off between parameter usage and computation steps.

\subsection{Decomposed CNN Modules and Query Policies}

\begin{figure}[h]
\centering
\begin{subfigure}[b]{0.45\textwidth}
\centering
\centering
\includegraphics[width=0.85\textwidth,height=6.5cm,keepaspectratio]{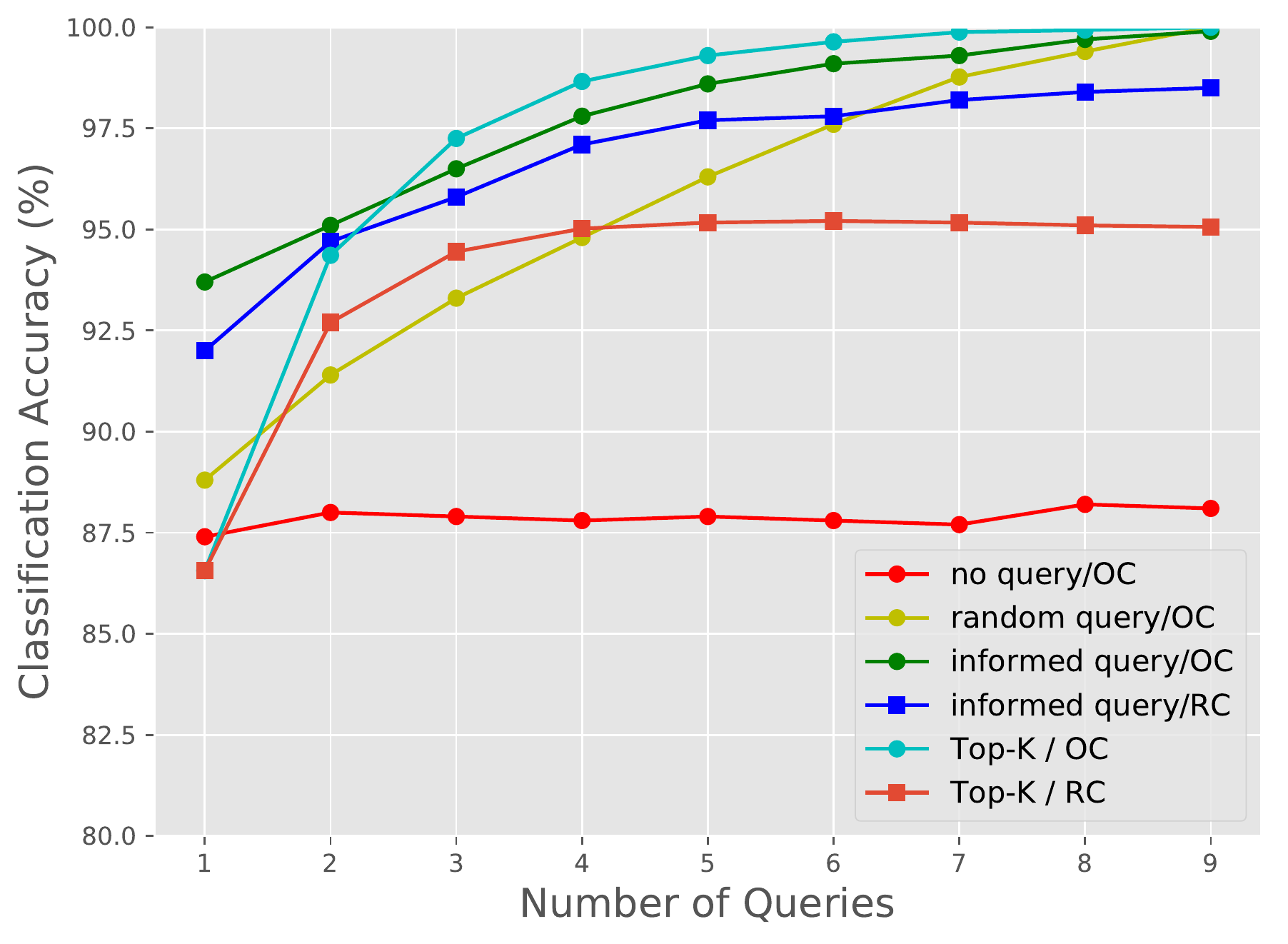}
\caption{\label{fig:decompose2}}
\end{subfigure} 
\begin{subfigure}[b]{0.45\textwidth}
\centering
\includegraphics[width=0.85\textwidth,height=8cm,keepaspectratio]{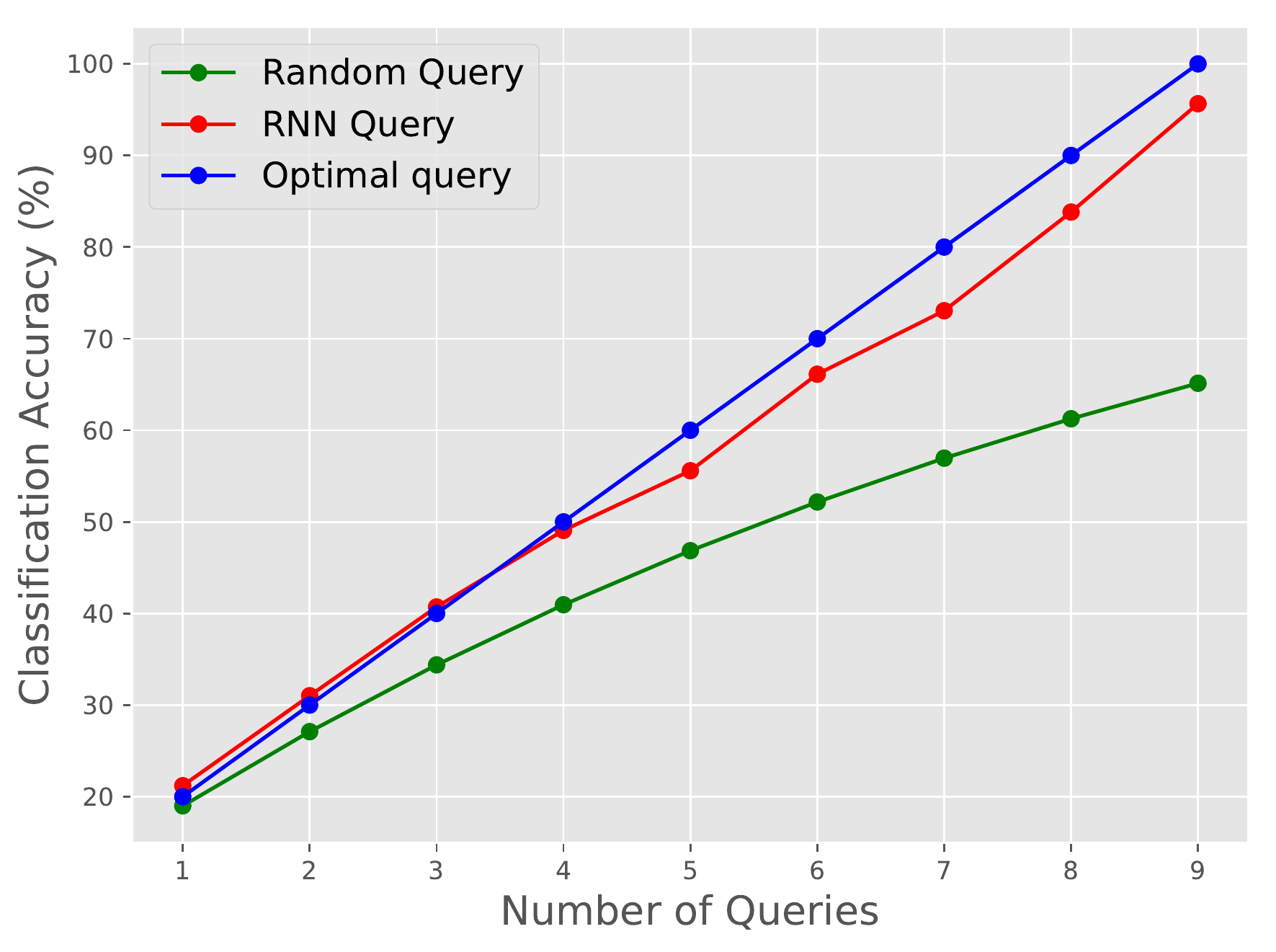}
\caption{\label{fig:decompose}}
\end{subfigure}
\caption{(a) Results from combining different recognition and comparator networks. OC indicates an oracle comparator while RC indicates a real (pre-trained network) comparator. Top-K indicates a query policy of performing verifications with all of the top-k predictions of the recognition network. (b) Results of RNN with oracle comparator. Random and optimal query results are theoretical rather than empirical.}
\end{figure}

Figure \ref{fig:decompose} shows the model's ability to classify when there is no recognition component present but only a verification component. In this case the comparator used is an oracle comparator, which allows us to isolate whether the RNN is actually capable of learning a reasonable query policy. The RNN's learned policy is compared against a random query policy and the optimal query policy (never repeating a query). From the figure we see that the model is able to conduct a better than random query policy, but is not able to achieve optimal performance. We also observed that performance increases with a higher RNN size up to 200. This suggests that the model is in some part able to track previously unsuccessful queries and remember if there was a match. However, its memory is not perfect and it cannot achieve optimal performance.


Figure \ref{fig:decompose2} reports the results when the output of a weak classifier (86.87\% accuracy) is concatenated with the result of the comparator. The weakest baselines is a no-query model which runs the RNN for the same number of steps as other models but does not receive information from a query. Stronger baselines include a random query policy with no repeats and a top-k query policy which queries all top-k classes from the recognition network. The informed-query model learns to use the predictions from the recognition network to output a query. For this model we experimented with both an oracle and a real (neural network) comparator.

Of the query-based models, informed queries performed the best over random and no query models. This demonstrates that the RNN controller is able to learn a policy that takes advantage of the recognition model's predictions and performs better than random. Between the oracle and real comparator models, there was small drop in performance as expected. Interestingly the informed model achieves greater performance up to two queries versus the top-k model. 

\section{Limitations and Future Work}

From the experiments we can conclude that the model can be pushed to learn a query-like behavior as originally hypothesized. However this only occurred in the case in which the recognition model and comparator models were separated rather than as a unified component. Simply concatenating channels was not a sufficient approach to encourage verification behavior. This suggests that a more appropriate pipeline for our model is to perform a recognition operation which is then followed by verification, rather than perform them simultaneously.

This new pipeline would also imply that our model may be less well suited for the one-shot learning task than initially believed, as a reasonably-well trained recognition module is required as the first step. Instead future work should focus on using this pattern of recognition-then-verification for challenging classifications such as datasets with similar looking images, or fine-tuning accuracy on standard datasets.

That said, the paradigm does have potential in the few-shot learning space, however would need to be tested on datasets with a larger number of classes such as mini-Imagenet \cite{Miniimagenet} or WebFace \cite{WebFace}. In this scenario, the recognition module would narrow the number of classes and the verification network would select the correct class. We could experiment with models and training regimes better suited for verification such as prototypical networks. In this way, we could extend few-shot learning to a large number of classes.

\section{Conclusion}

We demonstrated that it is possible to train a model to intelligently use both recognition and verifications capabilities to classify images. Notably, this was achieved using a recurrent-model with non-differentiable queries. This signals the potential for models to use non-differentiable components as aids. This opens the door for not only improved classification, but even for translation, question-answering, and any other challenging tasks in which external information or computation could provide a benefit. A key component of an intelligent agent is the ability to use tools, hence tasks of this form are pre-requisite if neural-network models are considered to be as such \cite{2001Space}.

\newpage
\bibliographystyle{abbrv} 
\bibliography{sample}
\end{document}